\documentclass[letterpaper]{article}
\usepackage[preprint]{aaai2027}
\usepackage[hyphens]{url}
\usepackage{graphicx}
\usepackage{natbib}
\usepackage{caption}
\usepackage{amsmath,amssymb,mathtools}
\usepackage{booktabs,multirow,array}
\usepackage{colortbl}

\definecolor{bestgreen}{RGB}{178,220,167}
\definecolor{secondgreen}{RGB}{220,237,216}
\newcommand{\bestcell}[1]{\cellcolor{bestgreen}\raisebox{-0.45pt}[0pt][0pt]{\textbf{#1}}}
\newcommand{\secondcell}[1]{\cellcolor{secondgreen}\raisebox{-0.45pt}[0pt][0pt]{#1}}
\newcommand{\bestcellfoot}[1]{\cellcolor{bestgreen}\raisebox{-0.55pt}[0pt][0pt]{\textbf{#1}}}
\newcommand{\secondcellfoot}[1]{\cellcolor{secondgreen}\raisebox{-0.55pt}[0pt][0pt]{#1}}

\newcommand{\method}{PIC-UIE}

\newcommand{\ours}[1]{\textbf{#1}}
\newcommand{\Fig}[1]{Figure~\ref{#1}}
\newcommand{\Tab}[1]{Table~\ref{#1}}

\newcommand{\figureplaceholder}[2]{%
  \fbox{\rule{0pt}{#2}%
  \rule{\dimexpr#1-2\fboxsep-2\fboxrule\relax}{0pt}}}
\newcommand{\figureasset}[3]{%
  \IfFileExists{#1}{\includegraphics[width=#2]{#1}}{\figureplaceholder{#2}{#3}}}

\title{PIC-UIE: Predicting Image-Adaptive Corrections for \\ Lightweight Underwater Image Enhancement}

\author{%
Cunhao Zhu\textsuperscript{\rm 1},
Dongliang Xu\textsuperscript{\rm 2},
Xiangtao Kong\textsuperscript{\rm 3},\\
Xiaoyan Lu\textsuperscript{\rm 4},
Tianyu Wang\textsuperscript{\rm 5},
Yue Yao\textsuperscript{\rm 1}\corresponding
}
\affiliations{%
\textsuperscript{\rm 1}Shandong University\\
\textsuperscript{\rm 2}School of Airspace Science and Engineering, Shandong University\\
\textsuperscript{\rm 3}The Hong Kong Polytechnic University\\
\textsuperscript{\rm 4}Shandong Tongyu Network Security Technology Co., Ltd.\\
\textsuperscript{\rm 5}Mohamed bin Zayed University of Artificial Intelligence
}

\begin{document}
\maketitle

\begin{abstract}
Underwater image enhancement (UIE) aims to restore visibility, color fidelity,
and structural detail from images degraded by wavelength-dependent attenuation
and backscatter. State-of-the-art UIE methods often rely on large backbones and
dense image-to-image prediction, limiting their practicality for edge
deployment. Moreover, operating entirely in a single color space couples
degradation estimation with luminance and chroma correction. To address these
challenges, we propose PIC-UIE, a lightweight predictor--executor framework
that predicts image-adaptive corrections from a fixed $256\times256$ RGB
thumbnail and applies them to the native-resolution input in the YCbCr color
space. The predictor produces seven outputs, organized into spatial correction,
nonlinear luminance and coupled chroma mapping, and image-level color
calibration. A depth map regularizes the transmission proxy during training,
whereas inference uses only the RGB input. With 9,486 parameters and 0.094
GFLOPs at $256\times256$, PIC-UIE achieves 24.137 dB PSNR and 0.9216 SSIM on
UIEB-90 and 21.320 dB PSNR on zero-shot LSUI. It further processes native 4K
images at 55.0 FPS under the comparison protocol. These results show that
structured correction prediction provides an effective and practical
alternative to dense RGB reconstruction for underwater image enhancement.
\end{abstract}

\section{Introduction}
\label{sec:intro}

Underwater images are degraded by wavelength-dependent attenuation and
backscatter, which vary with scene distance and water conditions
\citep{akkaynak2018revised,akkaynak2019seathru}.  The resulting color casts,
veiling haze, and contrast loss impair both visual inspection and downstream
perception.  Underwater image enhancement (UIE) aims to restore visibility and
color fidelity while preserving scene structure.  For embedded cameras and
robotic platforms, practical UIE systems must also process high-resolution
inputs within tight compute, memory, and latency budgets.

Recent learning-based UIE methods employ convolutional networks, Transformers,
state-space models, and diffusion models
\citep{islam2020funie,peng2023ushape,guan2024watermamba,zhao2024wfdiff}.
Despite strong restoration performance, many rely on large backbones,
multi-stage processing, or dense image-to-image prediction, making
native-resolution deployment costly on edge devices.  Moreover, using a single color space for both degradation estimation and correction overlooks the distinct strengths of different representations. RGB captures channel-wise attenuation cues, whereas YCbCr explicitly separates luminance from chroma. This motivates prediction in RGB and correction in YCbCr.

\begin{figure}[!t]
\centering
\figureasset{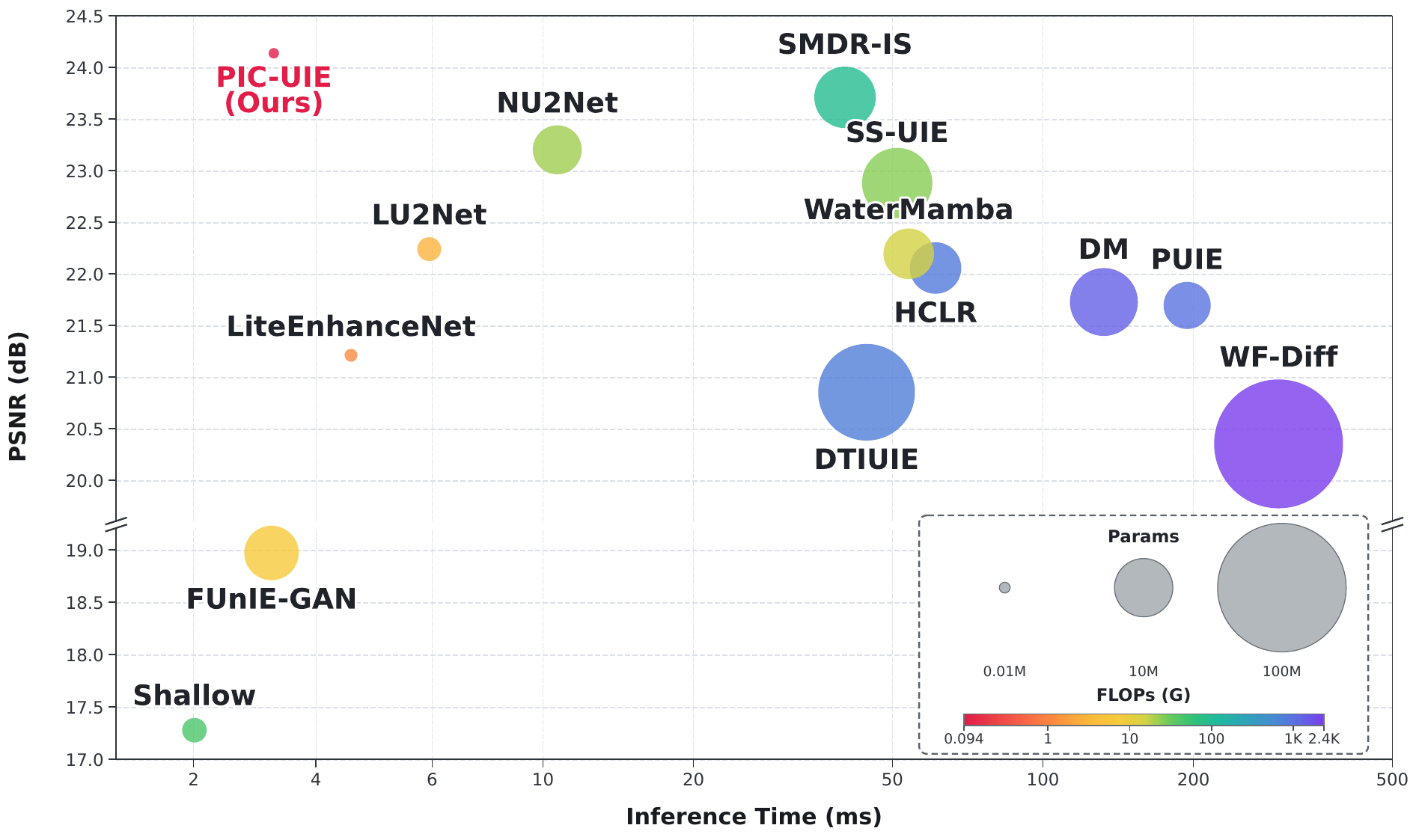}{\columnwidth}{4.1cm}
\caption{Quality--efficiency comparison on UIEB-90.  Bubble area indicates
parameter count, color indicates FLOPs, and the red marker denotes \method{}.}
\label{fig:quality_overview}
\end{figure}

We therefore propose \method{}, which combines a predictor with an executor.
The predictor estimates image-adaptive corrections from a fixed
$256\times256$ RGB thumbnail, while the executor applies them to the
native-resolution input in YCbCr.  This separation confines learned prediction
to a fixed resolution while preserving high-frequency structure in the input.
The predictor emits seven outputs organized into three groups: a transmission
proxy, veiling-light field, and luminance-gain field for spatial correction; a
monotone tone curve and coupled $9\times9$ chroma LUT for nonlinear mapping;
and a color correction matrix and covariance-based transform for image-level
calibration.  Together, these components define a compact correction space in
which the network predicts how to modify the input rather than reconstructing
every output pixel.  A depth map regularizes the transmission proxy during
training, while inference requires RGB alone.

We evaluate quality, transfer, speed, and task utility.  \mbox{\method{}} uses 9,486
parameters and
0.094 GFLOPs at $256\times256$, reaches 24.137 dB PSNR and 0.9216 SSIM on
UIEB-90 and 21.320 dB PSNR on zero-shot LSUI, and, under the comparison
  protocol, processes native 4K images at 55.0 FPS.  Matched-capacity output
  controls, representation choices, and physical diagnostics further isolate
  the effect of the structured correction space.  The resulting quality--efficiency trade-off is summarized
in \Fig{fig:quality_overview}.

The main contributions are:
\begin{itemize}
\item We recast lightweight UIE as predicting compact image-adaptive
  corrections followed by native-resolution execution, decoupling predictor
  cost from the input resolution.
\item We introduce an RGB predictor and YCbCr executor whose seven structured
  outputs cover spatial correction, nonlinear luminance--chroma mapping, and
  image-level color calibration, with depth regularization used only during
  training.\looseness=-1\par\looseness=0
\item Experiments across paired enhancement, zero-shot transfer,
  native-resolution inference, and downstream perception demonstrate a
  favorable quality--efficiency trade-off with only 9,486 parameters.
\end{itemize}

\section{Related Work}
\label{sec:related}

\paragraph{Learning-based UIE and color representations.}
UIEB provides a widely used paired real-image benchmark for data-driven UIE
\citep{li2019underwater}.  FUnIE-GAN emphasizes fast perceptual enhancement
\citep{islam2020funie}; later work considers predictive uncertainty
\citep{fu2022puie}, long-range interactions \citep{peng2023ushape}, and
multiscale detail \citep{zhang2024smdris}.  Other recent designs use
state-space models, frequency diffusion, dual-domain learning, or task-aware
supervision
\citep{guan2024watermamba,zhao2024wfdiff,peng2025adaptive,lin2026dtiuie}.
Color representation is another recurring choice.  UIEC\textsuperscript{2}-Net
encodes two color spaces \citep{wang2021uiec2}, whereas Ucolor combines several
color spaces under transmission guidance \citep{li2021ucolor}.  These methods
demonstrate the value of complementary color representations, but their
enhancement remains a dense image-to-image mapping at the processing resolution.

\paragraph{Image-adaptive operators and efficient UIE.}
A common strategy for resolution-efficient enhancement is to infer compact
transformation parameters from a thumbnail and apply the resulting operator
to the full-resolution image.  Prior work realizes this strategy using
bilateral grids, blended 3D LUTs, spatially varying lookup, and adaptive table
sampling
\citep{gharbi2017bilateral,zeng2020lut,wang2021spatiallut,yang2022adaint}.
These methods primarily target general photographic enhancement, so they
provide architectural context for our predictor--executor design rather than
task-matched UIE baselines.  Within UIE, efficiency is usually pursued through
compact end-to-end restoration networks such as FUnIE-GAN
\citep{islam2020funie}, which still predict enhanced pixels densely at their
processing resolution.  \method{} bridges these directions by combining
fixed-resolution parameter inference with a UIE-specific native-resolution
executor.  Its predictor estimates structured controls for attenuation,
veiling, luminance, coupled chroma, and image-level calibration.

\paragraph{Physical priors and depth supervision.}
Physics-based image-formation models describe attenuation and backscatter as
functions of wavelength and scene distance
\citep{akkaynak2018revised,akkaynak2019seathru}.  Learning-based systems use
geometry at different stages: Ucolor predicts transmission-guided features
\citep{li2021ucolor}, Osmosis conditions a diffusion prior on RGBD observations
\citep{bar2024osmosis}, and model-guided frameworks jointly optimize enhancement
and depth estimation \citep{du2024physical}.  Geometry is therefore supplied at
inference, predicted internally, or treated as an auxiliary output.  We use
relative depth only to regularize the transmission proxy during training.  The
deployed model remains RGB-only and contains no depth estimator or depth branch.

\section{Method}
\label{sec:method}

\begin{figure*}[!t]
\centering
\figureasset{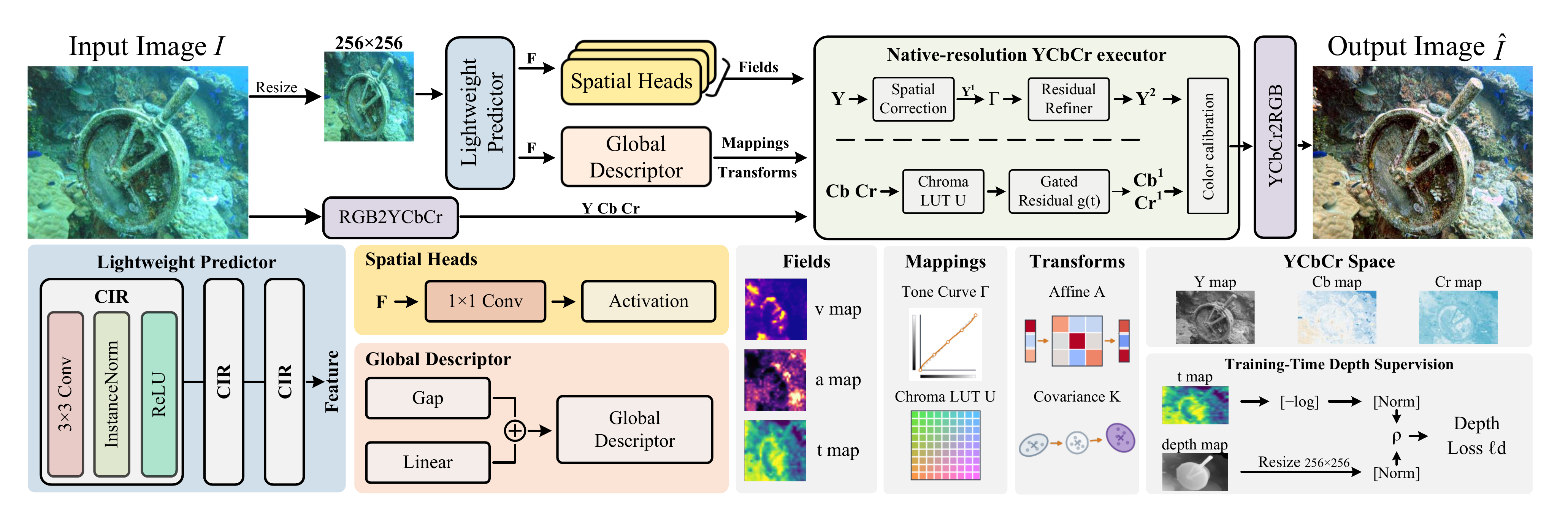}{\textwidth}{2.2cm}

\caption{Overview of \method{}.  The predictor estimates seven corrections
from a fixed $256\times256$ thumbnail, and the YCbCr executor applies them at
native resolution.  Depth supervises $t$ only during training.}
\label{fig:framework}
\end{figure*}

\subsection{Predictor--Executor Overview}
\label{sec:formulation}

\Fig{fig:framework} illustrates the overall predictor--executor framework.
Let $I\in[0,1]^{3\times H\times W}$ be an underwater RGB image of height $H$
and width $W$.  Let $\mathcal{R}_{256}$ denote bilinear resizing and
$\tilde I=\mathcal{R}_{256}(I)$ the resulting thumbnail.  We use
$\mu(\tilde I)$ and $\sigma(\tilde I)$ for its three-channel mean and standard
deviation, respectively, and collect the thumbnail statistics as
$s(\tilde I)=[\mu(\tilde I),\sigma(\tilde I)]$.  Let $P_\omega$ denote the
predictor with learned weights $\omega$, $\mathcal{T}$ the differentiable
RGB-to-YCbCr transform, and $\mathcal{E}$ the executor.  The predicted
correction set $\theta$ and enhanced image $\hat I$ are
\begin{equation}
\theta=P_\omega\!\left(\tilde I,s(\tilde I)\right),\qquad
\hat I=\mathcal{T}^{-1}\!\left(\mathcal{E}(\mathcal{T}(I);\theta)\right).
\label{eq:overview}
\end{equation}
The seven outputs are $\theta=\{t,v,a,\Gamma,U,A,K\}$: $t\in[0.05,1]$ is a
transmission proxy, $v\in[0,1]$ a veiling-light field, $a\geq0$ a luminance-gain
coefficient field, $\Gamma$ a monotone luminance curve, $U$ a coupled chroma
LUT, $A$ an affine color transform comprising a color correction matrix and
bias, and $K$ a covariance calibration transform.  These outputs form the
spatial, nonlinear, and calibration groups in \Fig{fig:framework}.
$P_\omega$ uses three $3\times3$ stride-2 convolutions of width 16, each
followed by affine InstanceNorm and ReLU.  The resulting $32\times32$ feature
map feeds the three spatial heads.  Global average pooling gives a
16-dimensional descriptor, to which a learned projection of $s(\tilde I)$ is
added before the nonspatial heads.  This explicit path retains image-level
color statistics that normalization may suppress.  Each head is initialized
near an input-preserving setting.  Prediction therefore remains fixed at
$256\times256$.  At native resolution, the only learned convolutional module
is a 465-parameter luminance refiner comprising a $1$-to-$16$ convolution, a
width-16 depthwise convolution, and a $16$-to-$1$ projection, all with
$3\times3$ kernels.\looseness=-1\par\looseness=0

\subsection{Structured YCbCr Corrections}
\label{sec:corrections}

Let $\mathcal{T}(I)=[Y,q]$, where $Y$ is luminance and
$q=(C_b,C_r)$ contains the blue- and red-difference chroma channels.  The
predictor receives RGB channel responses, while the executor uses YCbCr to
separate luminance processing from coupled chroma correction.

\paragraph{Spatial correction.}
We use $t$, $v$, and $a$ as learned proxies for spatial attenuation, veiling,
and luminance gain; the model does not estimate calibrated water coefficients.
After bilinear upsampling, the three fields produce the corrected luminance
$Y^{(1)}$.  Using
$\operatorname{clip}_{[0,1]}$ for elementwise clipping to $[0,1]$, we write
\begin{equation}
Y^{(1)}=\operatorname{clip}_{[0,1]}
\left([Y-(1-t)v]\,[1+a(1-t)]\right).
\label{eq:spatial}
\end{equation}
The first factor subtracts the predicted veil and the second applies a gain
that increases as $t$ decreases.  Clipping keeps luminance in its valid range.
The expression avoids division by small transmission values, which can amplify
noise in analytical inversion.

\paragraph{Nonlinear mapping.}
A shallow learned operator $F$ refines one-channel luminance, while a learned
gate $g$ adjusts chroma correction using $t$.  The stage produces luminance
$Y^{(2)}$ and chroma $q^{(1)}$ as
\begin{align}
Y^{(2)}&=\operatorname{clip}_{[0,1]}
 \left(\Gamma(Y^{(1)})+F(\Gamma(Y^{(1)}))\right), \nonumber\\
g(t)&=\operatorname{clip}_{[0,1]}\left(\beta+\gamma(1-t)\right), \nonumber\\
q^{(1)}&=q+g(t)\,[U(q)-q].
\label{eq:nonlinear}
\end{align}
$\Gamma$ remaps luminance globally, and the identity-centered
$9\times9$ table $U$ maps $C_b$ and $C_r$ jointly through bilinear
interpolation.  The learned global scalars $(\beta,\gamma)$ are initialized to
$(1,0)$; the gate can then use $t$ to vary the mapping strength across the
image, while the identity initialization of $U$ preserves the input initially.

\paragraph{Color calibration.}
Let $z^{(0)}=[Y^{(2)},q^{(1)}]$ be the corrected YCbCr representation, and let
$\mathcal{C}$ clip its luminance to $[0,1]$.  The affine transform $A=(M,b)$
performs cross-channel calibration.  For covariance calibration, let $m$ and
$S$ denote the per-image mean and regularized covariance of $z^{(1)}$, while
$u$, $G$, and $\alpha$ are the predicted target mean, lower-triangular
re-coloring matrix, and blend coefficient.  The two steps are
\begin{align}
z^{(1)}&=\mathcal{C}(Mz^{(0)}+b), \nonumber\\
K(z^{(1)})&=(1-\alpha)z^{(1)}
 +\alpha\left[G S^{-1/2}(z^{(1)}-m)+u\right], \nonumber\\
z^{(2)}&=\mathcal{C}(K(z^{(1)})).
\label{eq:calibration}
\end{align}
Here $\alpha$ is obtained from a sigmoid, and the $3\times3$ eigendecomposition
for $S^{-1/2}$ is computed once per image rather than per pixel.
The executor returns $\hat I=\mathcal{T}^{-1}(z^{(2)})$, converting back to RGB
only at the output.  Further head parameterizations and initialization details
are given in the supplementary material.

\subsection{Depth-Guided Training}
\label{sec:depthtraining}

Image reconstruction alone does not determine the semantics of $t$.  Let $d$
denote a training depth map.  Under the formation model, lower transmission is
associated with greater scene distance, so $-\log t$ should follow the relative
ordering in $d$.  We bilinearly resize $d$ to the spatial resolution of $t$
before evaluating the loss.  Let $\rho$ denote the per-image normalized
correlation obtained by standardizing both maps and averaging their
elementwise product.  The scale- and shift-invariant loss is
\begin{equation}
\ell_d=1-\rho\!\left(-\log t,d\right).
\label{eq:depth}
\end{equation}
The depth map appears only in this loss; it is not passed to $P_\omega$ or the
executor.  The full objective combines
RGB and red-channel reconstruction, SSIM and MS-SSIM structure terms
\citep{wang2004ssim,wang2003msssim}, the depth constraint, and smoothness and
monotonicity regularization for the chroma LUT.  Complete weights and training
details are provided in the supplementary material.  In the supplementary
loss ablation, removing any tested reconstruction or LUT term lowers UIEB PSNR
by 0.359--0.456 dB and LSUI PSNR by 0.036--0.173 dB; the two structural terms
produce the largest SSIM changes.

\begin{figure*}[!t]
\centering
\figureasset{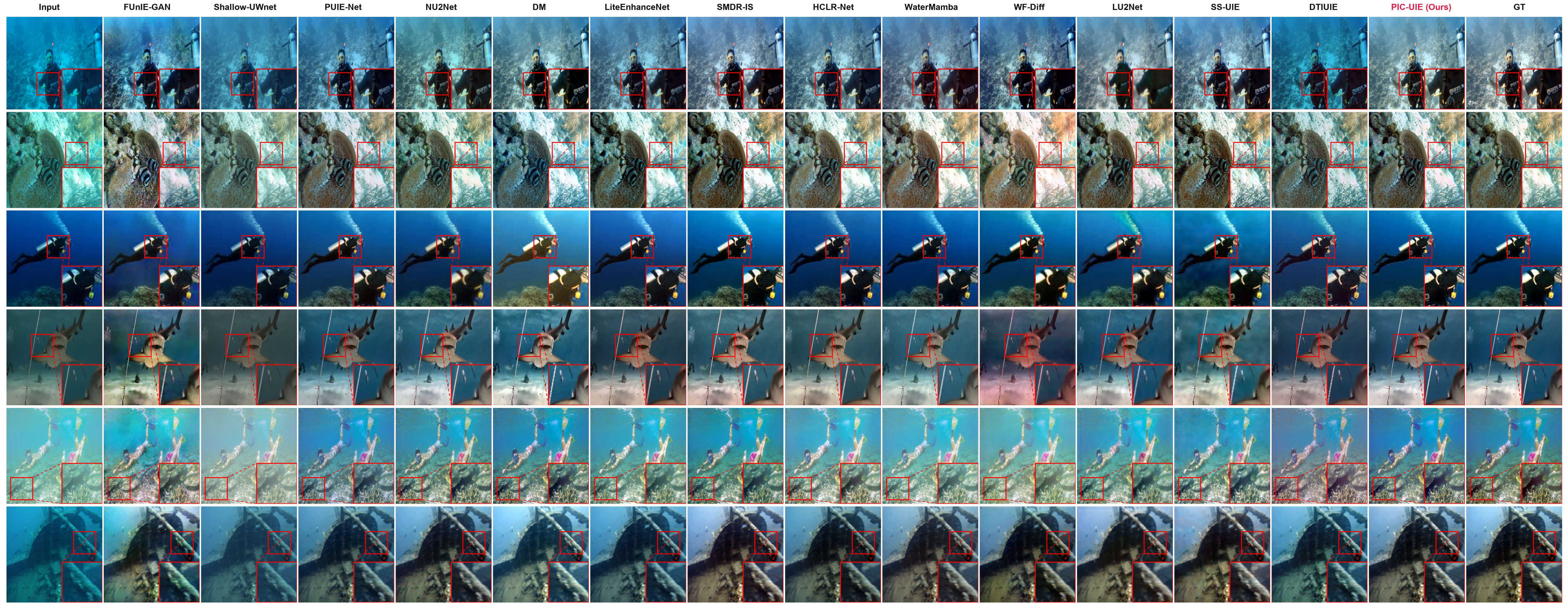}{\textwidth}{3.0cm}
\caption{Qualitative comparison on UIEB-90 and zero-shot LSUI.  Columns follow
the method order in \Tab{tab:main}; red boxes identify enlarged regions used to
compare color correction, local contrast, and structural-detail preservation.}
\label{fig:qualitative}
\end{figure*}

\begin{table*}[!t]
\centering
\begingroup
\scriptsize
\setlength{\tabcolsep}{1.5pt}
\renewcommand{\arraystretch}{0.90}
\begin{tabular*}{\textwidth}{@{\extracolsep{\fill}}l l c c c ccccc ccccc@{}}
\toprule
\multirow{2}{*}{Method} & \multirow{2}{*}{Venue}
 & \multirow{2}{*}{Params} & \multirow{2}{*}{FLOPs (G)}
 & \multirow{2}{*}{Time (ms)} & \multicolumn{5}{c}{UIEB-90}
 & \multicolumn{5}{c}{LSUI zero-shot}\\
\cmidrule(l{2pt}r{2pt}){6-10}\cmidrule(l{2pt}r{2pt}){11-15}
 & & & & & PSNR$\uparrow$ & SSIM$\uparrow$ & MSE$\downarrow$ & LPIPS$\downarrow$ & PCQI$\uparrow$
 & PSNR$\uparrow$ & SSIM$\uparrow$ & MSE$\downarrow$ & LPIPS$\downarrow$ & PCQI$\uparrow$\\
\midrule
FUnIE-GAN & RA-L'20 & 7.02M & 7.16 & \secondcell{3.11} & 18.969 & 0.7834 & 0.01506 & 0.2302 & 0.7472 & 18.058 & 0.7583 & 0.01763 & 0.2917 & \bestcell{0.7684}\\
Shallow-UWnet & AAAI'21 & 219K & 43.3 & \bestcell{2.01} & 17.279 & 0.7237 & 0.02383 & 0.2885 & 0.5212 & 20.311 & 0.8005 & 0.01344 & 0.2794 & 0.5961\\
PUIE-Net & ECCV'22 & 1.9M & 935.2 & 194.3 & 21.695 & 0.8941 & 0.00885 & 0.1259 & 0.6493 & 20.995 & 0.8249 & \bestcell{0.01005} & 0.2306 & 0.7076\\
NU\textsuperscript{2}Net & AAAI'23 & 3.15M & 20.8 & 10.68 & 23.202 & 0.9019 & 0.00644 & 0.1244 & 0.8164 & \bestcell{21.497} & 0.8261 & \secondcell{0.01007} & 0.2397 & 0.7216\\
DM-underwater & MM'23 & 18.3M & 1338 & 132.4 & 21.728 & \secondcell{0.9047} & 0.00928 & 0.1254 & \bestcell{0.8774} & 18.157 & 0.7757 & 0.02096 & 0.2765 & 0.7310\\
LiteEnhanceNet & ESWA'24 & \secondcell{13.7K} & \secondcell{1.18} & 4.52 & 21.212 & 0.8444 & 0.01035 & 0.1983 & 0.6974 & 20.282 & 0.8185 & 0.01253 & 0.2319 & 0.7334\\
SMDR-IS & AAAI'24 & 12.6M & 92.4 & 40.19 & \secondcell{23.710} & 0.8959 & \secondcell{0.00628} & \secondcell{0.1203} & 0.8125 & 20.609 & 0.8216 & 0.01163 & 0.2323 & 0.7097\\
HCLR-Net & IJCV'24 & 4.87M & 796.5 & 60.97 & 22.057 & 0.8663 & 0.00881 & 0.1415 & 0.7348 & 20.748 & \secondcell{0.8263} & 0.01102 & \bestcell{0.2193} & 0.7184\\
WaterMamba & arXiv'24 & 4.22M & 14.6 & 53.89 & 22.195 & 0.8631 & 0.00876 & 0.1426 & 0.7422 & 21.005 & \bestcell{0.8270} & 0.01073 & \secondcell{0.2220} & 0.7181\\
WF-Diff & CVPR'24 & 100.6M & 2375 & 296.0 & 20.354 & 0.8787 & 0.01311 & 0.1595 & 0.7743 & 19.779 & 0.7679 & 0.01307 & 0.2633 & 0.5828\\
LU2Net & arXiv'24 & 175.6K & 2.79 & 5.94 & 22.240 & 0.8169 & 0.00819 & 0.1932 & 0.6462 & 20.706 & 0.8039 & 0.01205 & 0.2451 & 0.7179\\
SS-UIE & AAAI'25 & 20.6M & 24.2 & 51.10 & 22.880 & 0.8870 & 0.00728 & 0.1519 & 0.8136 & 20.738 & 0.8132 & 0.01084 & 0.2377 & 0.7441\\
DTIUIE & TIP'26 & 50.8M & 728.8 & 44.38 & 20.854 & 0.8266 & 0.01102 & 0.1774 & 0.7542 & 20.534 & 0.8152 & 0.01268 & 0.2226 & 0.7182\\
\midrule
\ours{\method{}} & None & \bestcell{9.5K} & \bestcell{0.094} & 3.15 & \bestcell{24.137} & \bestcell{0.9216} & \bestcell{0.00606} & \bestcell{0.1060} & \secondcell{0.8732}
 & \secondcell{21.320} & 0.8213 & 0.01045 & 0.2266 & \secondcell{0.7666}\\
\bottomrule
\end{tabular*}
\endgroup
\caption{Full-reference results on UIEB-90 and zero-shot LSUI; dark and light
green cells denote the best and second-best results.  Baseline sources are
\citep{islam2020funie,naik2021shallow,fu2022puie,guo2023underwater,
tang2023diffusion,zhang2024liteenhance,zhang2024smdris,zhou2024hclr,
guan2024watermamba,zhao2024wfdiff,yang2024lu2net,peng2025adaptive,
lin2026dtiuie}.}
\label{tab:main}
\end{table*}

\begin{table*}[!t]
\centering
\begingroup
\footnotesize
\let\bestcell\bestcellfoot
\let\secondcell\secondcellfoot
\setlength{\tabcolsep}{2pt}
\renewcommand{\arraystretch}{0.90}
\begin{tabular*}{\textwidth}{@{\extracolsep{\fill}}ccccccccc@{}}
\toprule
\multirow{2}{*}{Method} & \multicolumn{4}{c}{U45 zero-shot} & \multicolumn{4}{c}{C60 zero-shot}\\
\cmidrule(l{2pt}r{2pt}){2-5}\cmidrule(l{2pt}r{2pt}){6-9}
 & UIQM$\uparrow$ & UCIQE$\uparrow$ & NIQE$\downarrow$ & Entropy$\uparrow$
 & UIQM$\uparrow$ & UCIQE$\uparrow$ & NIQE$\downarrow$ & Entropy$\uparrow$\\
\midrule
FUnIE-GAN & \bestcell{3.247} & 0.3057 & \bestcell{3.859} & 7.541 & \bestcell{3.143} & 0.2847 & \bestcell{5.313} & 7.167\\
Shallow-UWnet & 2.979 & 0.2334 & 4.722 & 6.880 & 2.433 & 0.2237 & 6.199 & 6.525\\
PUIE-Net & 3.172 & 0.2744 & 4.504 & 7.323 & 2.630 & 0.2651 & 5.789 & 7.086\\
NU\textsuperscript{2}Net & \secondcell{3.238} & 0.3040 & 4.311 & 7.526 & 2.808 & 0.2946 & 5.727 & 7.272\\
DM-underwater & 3.094 & 0.3229 & 4.274 & 7.513 & 2.675 & 0.3051 & 6.936 & 7.218\\
LiteEnhanceNet & 3.181 & 0.3079 & 4.373 & 7.459 & 2.581 & 0.2778 & 5.801 & 7.104\\
SMDR-IS & 3.150 & 0.3098 & 4.254 & 7.483 & 2.674 & \bestcell{0.3118} & \secondcell{5.563} & \secondcell{7.308}\\
HCLR-Net & 3.176 & 0.3082 & 4.305 & 7.498 & 2.810 & 0.2734 & 6.000 & 7.086\\
WaterMamba & 3.166 & 0.3112 & 4.738 & 7.507 & 2.826 & 0.3066 & 5.734 & \bestcell{7.313}\\
WF-Diff & 3.008 & 0.2986 & 4.750 & 7.370 & 2.797 & 0.2991 & 5.879 & 7.071\\
LU2Net & 3.123 & 0.3190 & 4.557 & 7.518 & 2.751 & 0.2941 & 5.872 & 7.230\\
SS-UIE & 3.061 & \bestcell{0.3294} & 4.605 & \bestcell{7.583} & 2.739 & 0.2882 & 5.954 & 7.116\\
DTIUIE & 3.167 & 0.2929 & \secondcell{4.074} & 7.298 & \secondcell{2.951} & 0.2791 & 6.586 & 7.063\\
\midrule
\ours{\method{}} & 3.120 & \secondcell{0.3251} & 4.241 & \secondcell{7.561}
 & 2.587 & \secondcell{0.3108} & 5.758 & 7.303\\
\bottomrule
\end{tabular*}
\endgroup
\caption{No-reference quality assessment on U45 and C60 under zero-shot
transfer at $256\times256$.  Higher UIQM, UCIQE, and entropy and lower NIQE
indicate better scores; dark and light green cells denote the best and
second-best values in each column, respectively.}
\label{tab:noref}
\end{table*}

\section{Experiments}
\label{sec:experiments}

\subsection{Experimental Setup and Protocol}
\label{sec:setup}

\paragraph{Data.}
Training uses UIEB-800, with UIEB-90 as its held-out paired test set
\citep{li2019underwater}; LSUI is evaluated zero-shot without fine-tuning
\citep{peng2023ushape}.  No-reference tests use U45 \citep{li2019fusion} and
UIEB's unpaired C60 set \citep{li2019underwater}.  UIQAD supplies native-resolution
images \citep{chu2023sisc}, while FLSea \citep{randall2026flsea} and URPC2019
\citep{urpc2019} support matching and detection, respectively.

\paragraph{Baselines.}
We compare the learning-based UIE methods listed in the tables.  Each baseline
was retrained on UIEB-800 using its official codebase.

\paragraph{Training.}
We train with AdamW for 700 epochs, using an initial learning rate of
$3\times10^{-4}$, weight decay $10^{-4}$, cosine annealing, batch size 16,
and EMA decay 0.999 \citep{loshchilov2019adamw}.  Paired images are resized to
$256\times256$ and augmented with independent horizontal and vertical flips
and a random rotation by a multiple of $90^\circ$.

\paragraph{Metrics and timing.}
For paired evaluation, predictions and references are resized identically to
$256\times256$; OpenCV-loaded BGR images are converted once to RGB before
PSNR, SSIM, MSE, LPIPS-Alex, and PCQI.  SSIM uses
\texttt{pytorch\_msssim}; LPIPS uses the official \texttt{lpips} package with
AlexNet and $[-1,1]$ inputs \citep{zhang2018lpips}.
No-reference evaluation reports UIQM, UCIQE, NIQE, and entropy; the first three
follow Panetta et al., Yang and Sowmya, and Mittal et al., respectively
\citep{panetta2016uiqm,yang2015uciqe,mittal2013niqe}.  Native throughput is
measured with batch size one and FP32 on a 48-GB NVIDIA GeForce RTX 4090.  We
use 10 warmup iterations followed by 30 timed iterations on random tensors and
exclude image loading.  FLOPs are recomputed with one profiler and counting
convention at the input resolution listed in each table; parameter counts use
the corresponding model configuration.

\subsection{Paired Quality Comparison}
\label{sec:maincompare}

\Tab{tab:main} reports the full-reference comparison on UIEB-90 and zero-shot
LSUI.  On UIEB-90, \method{} reaches 24.137 dB PSNR and 0.9216 SSIM, the
highest values in the table, and gives the lowest MSE and LPIPS.  Relative to
SMDR-IS, the differences are 0.427 dB in PSNR and 0.0257 in SSIM; MSE and LPIPS
are also lower.
On LSUI without fine-tuning, \method{} records 21.320 dB PSNR with 9.5K
parameters and 0.094 GFLOPs.  Its PSNR and PCQI rank second, whereas its SSIM,
MSE, and LPIPS are outside the top two.  The result is therefore strongest in
PSNR and contrast fidelity rather than uniform across all five metrics.

\Fig{fig:qualitative} presents representative paired comparisons from both
test sets using the same method ordering as \Tab{tab:main}.

\subsection{No-Reference Quality}
\label{sec:noref}

\Tab{tab:noref} evaluates images without paired references.  \method{} ranks
second in UCIQE and entropy on U45 and second in UCIQE on C60.

\subsection{High-Resolution Efficiency}
\label{sec:efficiency}

\begin{figure*}[!t]
\centering
\figureasset{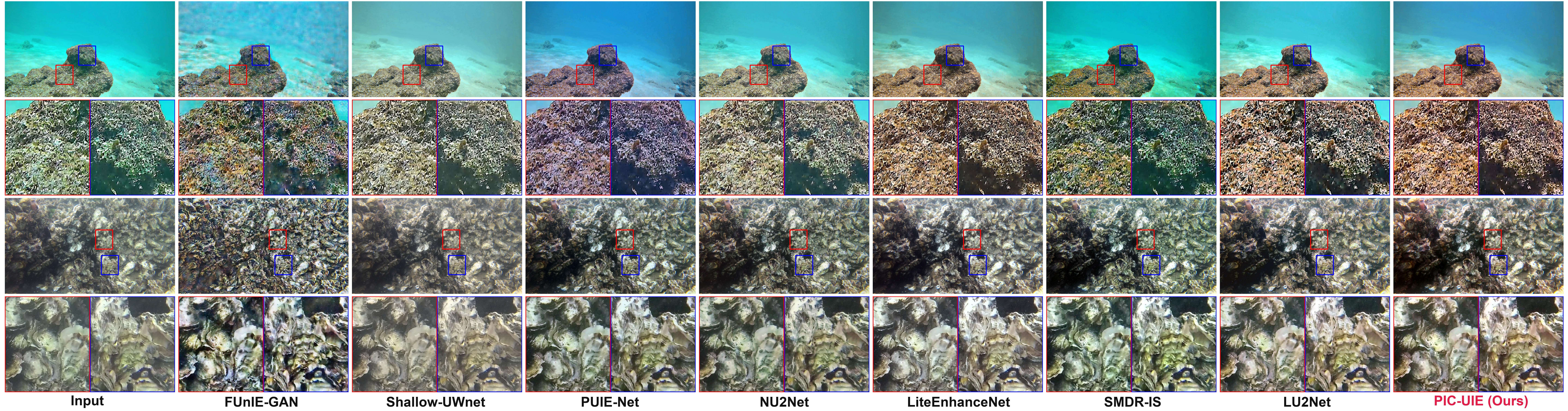}{\textwidth}{4.8cm}
\caption{Native 4K qualitative comparison on UIQAD.  Marked regions are shown
as enlarged crops to compare color recovery, local contrast, artifact
suppression, and fine-detail preservation across methods.}
\label{fig:hires4k}
\end{figure*}

The predictor always processes a $256\times256$ thumbnail.  Color conversion,
interpolation, LUT sampling, calibration, and the one-channel luminance refiner
account for the resolution-dependent cost.  Under the RTX 4090 protocol in
\Tab{tab:efficiency_downstream}, \method{} runs at 181.9 FPS at 1080P and 55.0
FPS at 4K, requiring 1.83 and 7.20 GFLOPs, respectively.  The 3.93$\times$
increase in FLOPs is close to the fourfold increase in pixel count.  Compared
with FUnIE-GAN, the fastest reported native-resolution baseline, \method{} is
2.34$\times$ faster at 1080P and 3.14$\times$ faster at 4K.  Its fixed-size
predictor requires 0.038 GFLOPs, or 2.1\% of total 1080P computation and 0.5\%
at 4K; most computation is therefore in the executor.  The no-reference
metrics show a trade-off: \method{} has the highest UCIQE at both resolutions
(0.3076 and 0.3203), while FUnIE-GAN has the highest UIQM.  In the examples in
\Fig{fig:hires4k}, the latter also introduces visible artifacts, so we do not
treat either metric alone as a perceptual ranking.\looseness=-1\par\looseness=0

\begin{table*}[!t]
\centering
\begingroup
\footnotesize
\let\bestcell\bestcellfoot
\let\secondcell\secondcellfoot
\setlength{\tabcolsep}{1pt}
\renewcommand{\arraystretch}{0.90}
\begin{tabular*}{\textwidth}{@{\extracolsep{\fill}}cccccccccccc@{\hspace{4pt}}}
\toprule
\multirow{2}{*}{Method} & \multicolumn{4}{c}{Native 1080P} & \multicolumn{4}{c}{Native 4K}
 & \multicolumn{2}{c}{FLSea} & URPC2019\\
\cmidrule(l{2pt}r{2pt}){2-5}\cmidrule(l{2pt}r{2pt}){6-9}\cmidrule(l{2pt}r{2pt}){10-11}\cmidrule(l{2pt}){12-12}
 & GFLOPs & FPS$\uparrow$ & UIQM$\uparrow$ & UCIQE$\uparrow$
 & GFLOPs & FPS$\uparrow$ & UIQM$\uparrow$ & UCIQE$\uparrow$
 & Matches$\uparrow$ & Inliers$\uparrow$ & mAP@50$\uparrow$\\
\midrule
FUnIE-GAN & 228.4 & \secondcell{77.6} & \bestcell{2.973} & 0.2503
 & 913.4 & \secondcell{17.5} & \bestcell{3.043} & 0.2630 & 1438 & 1359 & 0.723\\
Shallow-UWnet & 1369 & 24.4 & 2.076 & 0.2209
 & 5475 & 5.8 & 2.345 & 0.2335 & 665 & 630 & 0.725\\
PUIE-Net & 29589 & 0.6 & 2.419 & 0.2466
 & 118356 & 0.15 & 2.613 & 0.2779 & 1791 & 1706 & 0.745\\
NU\textsuperscript{2}Net & 662.0 & 29.7 & 2.524 & 0.2729
 & 2648 & 7.4 & 2.657 & 0.2756 & 2097 & 2007 & \bestcell{0.766}\\
DM-underwater & -- & -- & -- & -- & -- & -- & -- & -- & 574 & 544 & 0.742\\
LiteEnhanceNet & \secondcell{37.5} & 19.4 & 2.423 & 0.2751
 & \secondcell{150.0} & 4.6 & 2.601 & \secondcell{0.3053} & 1634 & 1551 & 0.748\\
SMDR-IS & 3117 & 5.6 & \secondcell{2.549} & 0.2806
 & 11775 & 0.9 & \secondcell{2.660} & 0.3009 & \secondcell{2181} & \secondcell{2071} & 0.740\\
HCLR-Net & 25176 & 1.6 & 2.435 & \secondcell{0.2825}
 & -- & -- & -- & -- & 1419 & 1350 & 0.737\\
WaterMamba & -- & -- & -- & -- & -- & -- & -- & -- & 549 & 509 & 0.734\\
WF-Diff & -- & -- & -- & -- & -- & -- & -- & -- & 249 & 226 & 0.749\\
LU2Net & 88.9 & 30.3 & 2.467 & 0.2793
 & 355.8 & 6.6 & 2.656 & 0.3041 & 1540 & 1463 & 0.717\\
SS-UIE & -- & -- & -- & -- & -- & -- & -- & -- & 356 & 327 & 0.721\\
DTIUIE & -- & -- & -- & -- & -- & -- & -- & -- & 420 & 391 & 0.737\\
\midrule
\ours{\method{}} & \bestcell{1.83} & \bestcell{181.9} & 2.502 & \bestcell{0.3076}
 & \bestcell{7.20} & \bestcell{55.0} & 2.638 & \bestcell{0.3203}
 & \bestcell{2185} & \bestcell{2078} & \secondcell{0.751}\\
\bottomrule
\end{tabular*}
\endgroup
\caption{Native-resolution efficiency, no-reference quality, and downstream
task performance.  GFLOPs and FPS use batch-one FP32 on an RTX 4090; FLSea
reports SIFT matches and RANSAC inliers, and URPC2019 reports YOLO11n mAP@50.
Dark and light green cells denote the best and second-best available results.}
\label{tab:efficiency_downstream}
\end{table*}

Dashes indicate fixed-size released evaluation pipelines or OOM under batch-one
FP32 on the 48-GB RTX 4090; method-specific cases are detailed in the supplement.

\subsection{Cross-Hardware Deployment}
\label{sec:cross_hardware}

We also measure \method{} beyond the RTX 4090, using a CPU, several desktop
GPUs, low-power inference accelerators, and an embedded Jetson platform.
\Tab{tab:cross_hardware} uses the same FP32 model on all six devices, without
architecture changes.  The 8-GB Jetson Orin NX reaches 29.5 FPS at native
1080P and 5.9 FPS at 4K.  The GTX 750 Ti processes 1080P images at 9.1 FPS,
while the Tesla T4 reaches 51.7 FPS at 1080P and 13.3 FPS at 4K.  Throughput
varies substantially with hardware, but neither reduced precision nor a
device-specific model is required for these measurements.

\begin{table}[!t]
\centering
\begingroup
\small
\setlength{\tabcolsep}{2.5pt}
\renewcommand{\arraystretch}{0.90}
\begin{tabular*}{\columnwidth}{@{\extracolsep{\fill}}lccc@{}}
\toprule
Hardware & Type / Memory & 1080P FPS$\uparrow$ & 4K FPS$\uparrow$\\
\midrule
Ryzen 7 4800H  & CPU                & 2.6  & 0.5\\
GTX 750 Ti     & GPU / 2 GB         & 9.1  & 2.2\\
Tesla P4       & GPU / 8 GB         & 23.7 & 6.0\\
Jetson Orin NX & Edge GPU / 8 GB    & 29.5 & 5.9\\
RTX 2060       & GPU / 6 GB         & 41.0 & 10.7\\
Tesla T4       & GPU / 16 GB        & 51.7 & 13.3\\
\bottomrule
\end{tabular*}
\endgroup
\caption{Cross-hardware FP32 throughput.}
\label{tab:cross_hardware}
\end{table}

The Orin NX is operated in MAXN mode with JetPack 6.2 (L4T R36.5.0), CUDA
12.6, and PyTorch 2.8.0.  Its reported values are the mean of 50 forward
passes after warm-up, timed with synchronized CUDA events.  Inputs are already
resident on the GPU before timing.\looseness=-1\par\looseness=0

\subsection{Downstream Task Evaluation}
\label{sec:downstream}

We next compare enhancement front-ends for geometry and recognition.
On FLSea, we match adjacent frames with SIFT and reject outliers with RANSAC
\citep{lowe2004sift,fischler1981ransac}.
On URPC2019, a separate YOLO11n detector is trained for each enhanced set under
the same optimization and evaluation protocol.
\Tab{tab:efficiency_downstream} reports the resulting averages.  \method{}
obtains 2,185 matches and 2,078 inliers on FLSea, ranking first for both
metrics; 95.1\% of its matches remain after RANSAC.  On URPC2019, it reaches
0.751 mAP@50, second to NU\textsuperscript{2}Net at 0.766.  The two tests place
\method{} near the top for both local matching and detection.

\subsection{Structured Design and Capacity}
\label{sec:design_capacity}

The matched-capacity controls in \Tab{tab:direct} change the output
parameterization while keeping the training recipe and parameter budget nearly
fixed.  Dense RGB and YCbCr residual predictors obtain 20.940 and 20.880 dB,
whereas the structured executor reaches 24.137 dB with 59 additional
parameters.  The two dense predictors differ by only 0.060 dB, compared with a
3.197--3.257 dB gap to structured correction.  Under this control, neither the
small parameter difference nor a direct switch to YCbCr accounts for the gain.

\begin{table}[!t]
\centering
\begingroup
\small
\setlength{\tabcolsep}{3pt}
\renewcommand{\arraystretch}{0.90}
\begin{tabular*}{\columnwidth}{@{\extracolsep{\fill}}lcc@{}}
\toprule
Output parameterization & Params & UIEB PSNR\\
\midrule
Direct RGB residual & 9,427 & 20.940\\
Direct YCbCr residual & 9,427 & 20.880\\
\ours{Structured corrections} & \ours{9,486} & \ours{24.137}\\
\bottomrule
\end{tabular*}
\endgroup
\caption{Matched-capacity output parameterizations.}
\label{tab:direct}
\end{table}

\Tab{tab:representation} varies the predictor input, chroma parameterization,
and treatment of global statistics separately.  These tests distinguish the
representation used to estimate degradation from the correction applied by the
executor.

RGB is the best-performing predictor input on both datasets.  It exceeds YCbCr
and RGB+YCbCr, supporting the use of RGB for prediction and YCbCr for execution
in this model.
The coupled 2D LUT also exceeds two independent 1D curves and a global affine
map.  Removing projected RGB statistics causes the largest drop in this table
(0.551 dB on UIEB and 0.256 dB on LSUI), while direct concatenation adds 1,184
parameters and remains below the full model.  Every tested replacement lowers
PSNR on both datasets, although the margins differ across representation
choices.

\begin{figure}[!t]
\centering
\figureasset{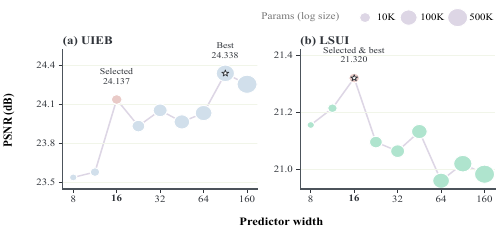}{0.92\columnwidth}{4.5cm}
\caption{Predictor-capacity scan on UIEB and LSUI.}
\label{fig:capacity}
\end{figure}

\paragraph{Capacity scaling.}
\Fig{fig:capacity} reports a post-hoc predictor-capacity diagnostic.  The final
architecture and hyperparameters, including width 16, were frozen before
evaluation on UIEB-90 and LSUI; these sweeps were not used for model selection.
Widths 8 and 12 trail width 16 on both datasets.  Width 96 improves UIEB by
0.201 dB but uses 20.2$\times$ more parameters and loses 0.301 dB on LSUI;
width 160 uses 53.1$\times$ more parameters for a 0.117 dB UIEB gain and a
0.338 dB LSUI drop.  Every tested width from 24 to 160 is below width 16 on
LSUI.

\begin{table}[!t]
\centering
\begingroup
\small
\setlength{\tabcolsep}{1.5pt}
\renewcommand{\arraystretch}{0.90}
\begin{tabular*}{\columnwidth}{@{\extracolsep{\fill}}lccc@{}}
\toprule
Variant & Params & UIEB & LSUI\\
\midrule
Predictor: YCbCr input & 9,486 & 23.837 & 21.178\\
Predictor: RGB+YCbCr & 9,918 & 23.766 & 21.241\\
Chroma: two 1D curves & 7,276 & 23.869 & 21.239\\
Chroma: global affine & 6,834 & 23.721 & 21.260\\
w/o RGB statistics & 9,374 & 23.586 & 21.064\\
Statistics: concatenation & 10,670 & 23.781 & 21.196\\
\ours{Full model} & \ours{9,486} & \ours{24.137} & \ours{21.320}\\
\bottomrule
\end{tabular*}
\endgroup
\caption{Representation choices (PSNR).}
\label{tab:representation}
\end{table}

\subsection{Component and Depth Diagnostics}
\label{sec:component_depth}

\Tab{tab:ablation} tests executor components with independently trained
variants.  Removing covariance calibration and the CCM reduces UIEB/LSUI by
0.910/0.456 and 0.730/0.300 dB, respectively.  Removing the physical luminance
transform, tone curve, chroma LUT, or transmission gate also lowers PSNR on
both datasets, but by smaller margins.  The two calibration components account
for the largest changes in this ablation.

\begin{table}[!t]
\centering
\begingroup
\small
\setlength{\tabcolsep}{2pt}
\renewcommand{\arraystretch}{0.90}
\begin{tabular*}{\columnwidth}{@{\extracolsep{\fill}}lcc@{}}
\toprule
Variant & UIEB PSNR & LSUI PSNR\\
\midrule
w/o physical luminance & 23.700 & 21.259\\
w/o tone curve & 23.745 & 21.219\\
LUT $\rightarrow$ affine & 23.721 & 21.260\\
w/o CCM & 23.407 & 21.020\\
w/o covariance calibration & 23.227 & 20.864\\
w/o transmission gate & 23.697 & 21.242\\
w/o depth supervision & 23.897 & 21.181\\
\ours{Full model} & \ours{24.137} & \ours{21.320}\\
\bottomrule
\end{tabular*}
\endgroup
\caption{Leave-one-out ablation (PSNR).}
\label{tab:ablation}
\end{table}

\paragraph{Depth supervision.}
Removing depth supervision lowers UIEB-90 from 24.137 to 23.897 dB and LSUI
from 21.320 to 21.181 dB.  Depth enters only the training loss, so the 0.240
and 0.139 dB differences add no depth input, inference branch, parameters, or
deployment-time computation.  Shuffled depth targets and spatial random noise
retain the correlation loss but break input alignment; they obtain 23.934 and
23.845 dB on UIEB-90.  Their gap to the aligned target is consistent with a
benefit from relative depth ordering rather than from adding an arbitrary
auxiliary objective.

\begin{figure}[!t]
\centering
\figureasset{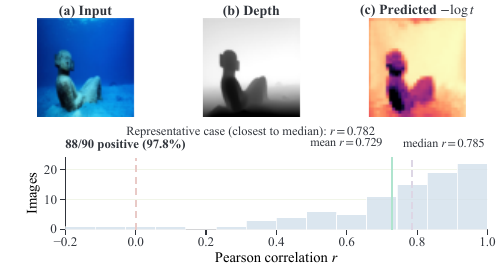}{0.92\columnwidth}{4.2cm}
\caption{Depth-consistency diagnostic on UIEB-90.}
\label{fig:depth}
\end{figure}

In \Fig{fig:depth}, the per-image correlation between $-\log t$ and depth has
mean 0.729 and median 0.785, and is positive on 88 of 90 images.  Without depth
supervision, the mean is 0.179.  The diagnostic supports relative ordering of
the transmission proxy; it does not imply metric-depth recovery.

\paragraph{Overall interpretation.}
\Tab{tab:representation} changes correction representation, whereas
\Tab{tab:ablation} removes executor components under independent training.
All alternatives lower UIEB and LSUI PSNR; calibration and RGB-statistics
injection produce the largest changes.  Matched dense regressors trail
structured correction by 3.197--3.257 dB at similar parameter counts, while
20.2$\times$--53.1$\times$ wider predictors gain at most 0.201 dB on UIEB and
lose at least 0.301 dB on LSUI.  Thus, the structured output space offers the
strongest joint trade-off among the tested configurations. Together, the matched-capacity and width controls separate representation from
scale.  The former change what the predictor emits at nearly fixed capacity,
whereas the latter increase capacity without altering the executor.  Their
contrasting trends indicate that the gain comes from the structure of the
correction space rather than parameter count alone.  Additional predictor
parameters are therefore not an effective substitute for structured outputs.

\section{Conclusion}
\label{sec:conclusion}

We introduced \method{}, a predictor--executor with 9,486 parameters that
estimates corrections from a fixed thumbnail and applies them at native
resolution in YCbCr.  It combines spatial luminance correction, a coupled chroma LUT, and
image-level calibration, with relative depth used only during training.  At
matched capacity, this structured output space improves UIEB PSNR by
3.197--3.257 dB over dense RGB and YCbCr residual prediction, reaching 24.137
dB on UIEB-90 and 21.320 dB on zero-shot LSUI.  In FP32, \method{} runs at 55.0
FPS on native 4K inputs on an RTX 4090 and at 29.5 FPS at 1080P on an 8-GB
Jetson Orin NX; it ranks first for FLSea matching and second for URPC2019
detection.  These results show that fixed-size structured prediction scales
effectively across desktop and edge hardware.

\bibliography{references}

\end{document}


\maketitle

\section{Exact Model Specification}

\subsection{Predictor}

The predictor receives a bilinearly resized $256\times256$ RGB thumbnail.  Its
body consists of three $3\times3$ stride-2 convolutions of width 16, each
followed by affine InstanceNorm and ReLU.  The resulting $32\times32$ feature
map feeds three spatial heads.  Global average pooling produces a
16-dimensional descriptor for the remaining heads.  A $6$-to-$16$ linear
layer projects the RGB channel means and standard deviations and adds them to
this descriptor, keeping image-level color statistics available after
normalization.

\begin{table}[!ht]
\centering
\scriptsize
\setlength{\tabcolsep}{2pt}
\renewcommand{\arraystretch}{0.8}
\begin{tabular}{lcl}
\toprule
Head & Output size & Parameterization\\
\midrule
Transmission $t$ & $1\times32\times32$ & sigmoid, clamp $[0.05,1]$\\
Veiling field $v$ & $1\times32\times32$ & sigmoid\\
Gain field $a$ & $1\times32\times32$ & softplus\\
Tone curve $\Gamma$ & 32 increments & positive cumulative curve\\
Chroma LUT $U$ & $2\times9\times9$ & identity table plus residual\\
Affine transform $A$ & 12 values & $3\times3$ matrix plus bias\\
Covariance transform $K$ & 10 values & mean, lower triangle, blend\\
\bottomrule
\end{tabular}
\caption{The seven prediction heads in the released configuration.}
\label{tab:heads}
\end{table}

The tone head predicts 32 positive increments.  Cumulative normalization and a
zero endpoint yield 33 monotone samples.  The covariance head predicts three
target means, six entries of a lower-triangular matrix, and one blend logit.

\subsection{Parameter count}

The convolution and InstanceNorm layers contain 5,184 parameters.  RGB
statistics injection adds 112, the three spatial heads add 51, and the tone,
chroma-LUT, affine-color, and covariance heads add 544, 2,754, 204, and 170,
respectively.  This gives 9,019 predictor parameters.  Two chroma-gate scalars
and a 465-parameter luminance refiner bring the total to 9,486.  The refiner is
a $1$-to-$16$ convolution, a width-16 depthwise convolution, and a
$16$-to-$1$ projection, all with $3\times3$ kernels.  It operates on one
luminance channel rather than a native-resolution RGB feature pyramid.

\newpage
\subsection{Identity initialization}

The correction heads are initialized close to an input-preserving
configuration.  Transmission weights are zero with bias 4; veil and gain
weights are zero with biases $-4$; tone logits are zero; the chroma table and
affine matrix start from identity, with zero affine bias; and the covariance
blend starts near zero.  The statistics projection and the final
luminance-refiner projection are initialized as no-ops.  This configuration is
close to, but not exactly, identity because sigmoid and softplus outputs remain
finite.

\section{Executor and Training Details}

\subsection{Color conversion}

The executor uses differentiable full-range BT.601 with centered chroma:
\begin{align}
Y &= 0.299R+0.587G+0.114B,\nonumber\\
C_b &= 0.564(B-Y),\qquad C_r=0.713(R-Y).
\end{align}
The inverse conversion is
\begin{align}
R &= Y+1.402C_r,\qquad B=Y+1.773C_b,\nonumber\\
G &= \left(Y-0.299R-0.114B\right)/0.587.
\end{align}
RGB is clipped to $[0,1]$ after the final inverse conversion.

\subsection{Forward sequence}

The $32\times32$ fields $t$, $v$, and $a$ are bilinearly upsampled to the
native input size.  Spatial luminance correction is
\begin{equation}
Y^{(1)}=\operatorname{clip}_{[0,1]}
\left([Y-(1-t)v][1+a(1-t)]\right).
\label{eq:suppspatial}
\end{equation}
This bounded form avoids division by small transmission values.  Let $F$ be
the shallow luminance refiner and $q=(C_b,C_r)$.  The nonlinear stage is
\begin{align}
Y^{(2)} &= \operatorname{clip}_{[0,1]}
\left(\Gamma(Y^{(1)})+F(\Gamma(Y^{(1)}))\right),\nonumber\\
g(t) &= \operatorname{clip}_{[0,1]}\left(\beta+\gamma(1-t)\right),\nonumber\\
q^{(1)} &= q+g(t)[U(q)-q],
\label{eq:suppnonlinear}
\end{align}
where $U$ is queried by bilinear interpolation and $(\beta,\gamma)$ are learned
global scalars initialized to $(1,0)$.

Let $z^{(0)}=[Y^{(2)},q^{(1)}]$, and let $\mathcal C$ clip only the luminance
channel.  The predicted affine transform contains a matrix $M$ and bias $b$:
\begin{equation}
z^{(1)}=\mathcal C(Mz^{(0)}+b).
\end{equation}
For covariance calibration, let $m$ and $S$ be the per-image mean and
regularized covariance of $z^{(1)}$, $u$ the predicted target mean, $G$ the
predicted lower-triangular re-coloring matrix, and $\alpha$ the sigmoid of the
blend logit.  The final calibrated representation is
\begin{align}
K(z^{(1)})={}&(1-\alpha)z^{(1)}\nonumber\\
&+\alpha\left[G S^{-1/2}(z^{(1)}-m)+u\right],\nonumber\\
z^{(2)}={}&\mathcal C(K(z^{(1)})).
\label{eq:suppcov}
\end{align}
The $3\times3$ eigendecomposition used to obtain $S^{-1/2}$ is performed once
per image, not per pixel.  The executor converts $z^{(2)}$ back to RGB only at
the output.

\subsection{Objective and optimization}

Let $\mathcal L_1$ be RGB L1 loss, $\mathcal L_R$ red-channel L1 loss,
$\ell_d$ the depth constraint, and $\mathcal L_S$ and $\mathcal L_M$ the LUT
smoothness and monotonicity terms.  The complete objective is
\begin{align}
\mathcal L={}&\mathcal L_1+0.3\mathcal L_R
+0.6(1-\operatorname{SSIM})\nonumber\\
&+0.4(1-\operatorname{MS\text{-}SSIM})+0.1\ell_d
+0.05\mathcal L_S+\mathcal L_M.
\label{eq:objective}
\end{align}
$\mathcal L_S$ sums squared finite differences along both LUT axes.
$\mathcal L_M$ penalizes negative derivatives of the $C_b$ output along the
$C_b$ axis and of the $C_r$ output along the $C_r$ axis.

For a relative training depth map $d$, the depth loss is
\begin{equation}
\ell_d=1-\rho(-\log(\operatorname{clip}(t,10^{-3},1)),d),
\end{equation}
where both maps are standardized per image before their normalized correlation
$\rho$ is computed.  The loss is invariant to affine changes in depth scale.
The depth map is never concatenated with RGB and is not required at inference.

Training uses the 800 UIEB pairs for 700 epochs at $256\times256$.  We use
AdamW with batch size 16, initial learning rate $3\times10^{-4}$, weight decay
$10^{-4}$, cosine annealing, and EMA decay 0.999
\citep{loshchilov2019adamw}.  Augmentation consists of independent horizontal
and vertical flips and rotations by multiples of $90^\circ$.

\subsection{Evaluation protocol}

UIEB-90 is the held-out paired test set \citep{li2019underwater}; LSUI is used
for paired zero-shot evaluation without fine-tuning \citep{peng2023ushape}.
U45 \citep{li2019fusion} and C60 \citep{li2019underwater} are used only for
no-reference evaluation, and UIQAD supplies the native-resolution images
\citep{chu2023sisc}.  Full-reference predictions and targets are resized
identically to $256\times256$ before PSNR, SSIM, MSE, LPIPS-Alex, and PCQI are
computed.  OpenCV-loaded BGR images are converted once to RGB before metric
computation.  SSIM uses \texttt{pytorch\_msssim} with data range one and an
11-pixel Gaussian window \citep{wang2004ssim}; LPIPS uses the official
\texttt{lpips} package, AlexNet backbone, and $[-1,1]$ inputs
\citep{zhang2018lpips}.  The same resizing and metric implementations are used
for every method.

Runtime uses batch size one and FP32 on one 48-GB NVIDIA GeForce RTX 4090.  Each
measurement has 10 warmup iterations and 30 synchronized timed iterations on
random tensors; image loading is excluded.  All FLOPs are recomputed with the
same profiler and counting convention.  UIEB-90, LSUI, U45, and C60 use
$256\times256$ inputs, while the 1080P and 4K tests use their native resolutions.
In the native-resolution comparison, a dash denotes an unavailable measurement.
The released evaluation pipelines for DM-underwater, SS-UIE, and DTIUIE resize
inputs to $256\times256$, so native-resolution measurements are not reported.
WaterMamba and WF-Diff cannot complete either native-resolution test within
48 GB, while HCLR-Net exceeds memory at 4K.

\FloatBarrier
\section{Extended Ablations}

\subsection{Thumbnail and LUT resolution}

\begin{table}[!ht]
\centering
\scriptsize
\setlength{\tabcolsep}{2pt}
\begin{tabular}{rccc}
\toprule
Thumbnail & UIEB PSNR/SSIM & LSUI PSNR/SSIM & Pred. GFLOPs\\
\midrule
128 & 23.485/0.9185 & 21.119/0.8186 & 0.0095\\
192 & 23.700/0.9195 & 21.145/0.8192 & 0.0213\\
\ours{256} & \ours{24.137/0.9216} & \ours{21.320/0.8213} & 0.0379\\
384 & 23.872/0.9201 & 21.226/0.8197 & 0.0852\\
\bottomrule
\end{tabular}
\caption{Thumbnail-resolution diagnostic.}
\label{tab:thumbnail}
\end{table}

\begin{table}[!ht]
\centering
\small
\setlength{\tabcolsep}{3pt}
\begin{tabular}{rccc}
\toprule
LUT grid & Params & UIEB PSNR/SSIM & LSUI PSNR/SSIM\\
\midrule
$5\times5$ & 7,582 & 23.798/0.9207 & 21.194/0.8184\\
$7\times7$ & 8,398 & 23.816/0.9201 & 21.217/0.8189\\
\ours{$9\times9$} & \ours{9,486} & \ours{24.137/0.9216} & \ours{21.320/0.8213}\\
$11\times11$ & 10,846 & 23.785/0.9191 & 21.233/0.8201\\
\bottomrule
\end{tabular}
\caption{Coupled chroma-LUT grid-size diagnostic.}
\label{tab:lutgrid}
\end{table}

These sweeps are post-hoc diagnostics performed after the final configuration
was frozen; no UIEB-90 or LSUI result was used for model selection.  Among the
tested settings, the $256\times256$ thumbnail and $9\times9$ LUT give the
highest PSNR and SSIM on both datasets.  Increasing the thumbnail to 384 or the
LUT to $11\times11$ adds computation or parameters but lowers at least one
metric on each test set.

\subsection{Loss terms}

\begin{table}[!ht]
\centering
\scriptsize
\setlength{\tabcolsep}{3pt}
\begin{tabular}{lcc}
\toprule
Variant & UIEB PSNR/SSIM & LSUI PSNR/SSIM\\
\midrule
\ours{Full model} & \ours{24.137/0.9216} & \ours{21.320/0.8213}\\
w/o red-weighted L1 & 23.681/0.9201 & 21.284/0.8207\\
w/o SSIM & 23.778/0.9183 & 21.147/0.8169\\
w/o MS-SSIM & 23.713/0.9192 & 21.228/0.8200\\
w/o SSIM and MS-SSIM & 23.722/0.9124 & 21.215/0.8189\\
w/o LUT smoothness & 23.692/0.9202 & 21.260/0.8202\\
w/o LUT monotonicity & 23.711/0.9205 & 21.222/0.8197\\
\bottomrule
\end{tabular}
\caption{Loss-term ablation under the common training and evaluation
protocol.}
\label{tab:loss}
\end{table}

Removing SSIM or MS-SSIM produces the largest SSIM changes.  The red-weighted
term and both LUT regularizers mainly affect reconstruction PSNR, with smaller
changes in SSIM.

\section{Downstream Protocols}

\paragraph{FLSea matching.}
For each of five FLSea scenes \citep{randall2026flsea}, four frames around the
middle of the sequence are selected with stride four.  SIFT descriptors \citep{lowe2004sift} are matched
with a Lowe ratio of 0.75.  A homography is estimated with a 3-pixel RANSAC
threshold \citep{fischler1981ransac}; the main paper reports the mean number of
accepted matches and inliers over adjacent pairs.

\paragraph{URPC2019 detection.}
Each enhancement method is applied to the URPC2019 detector data
\citep{urpc2019} before training.  A
separate YOLO11n detector is trained for each enhanced set for 300 epochs with
patience 100, and the best validation checkpoint is evaluated on the common
test split.  This prevents a detector trained on one enhancement distribution
from being evaluated on another.

\bibliography{reference-supp}
\clearpage

\begin{figure*}[!t]
\section{Complete Qualitative Results}
\noindent The following figures present the complete unpaired,
native-resolution, and downstream comparison grids.
\vspace{6pt}
\centering
\includegraphics[width=\textwidth]{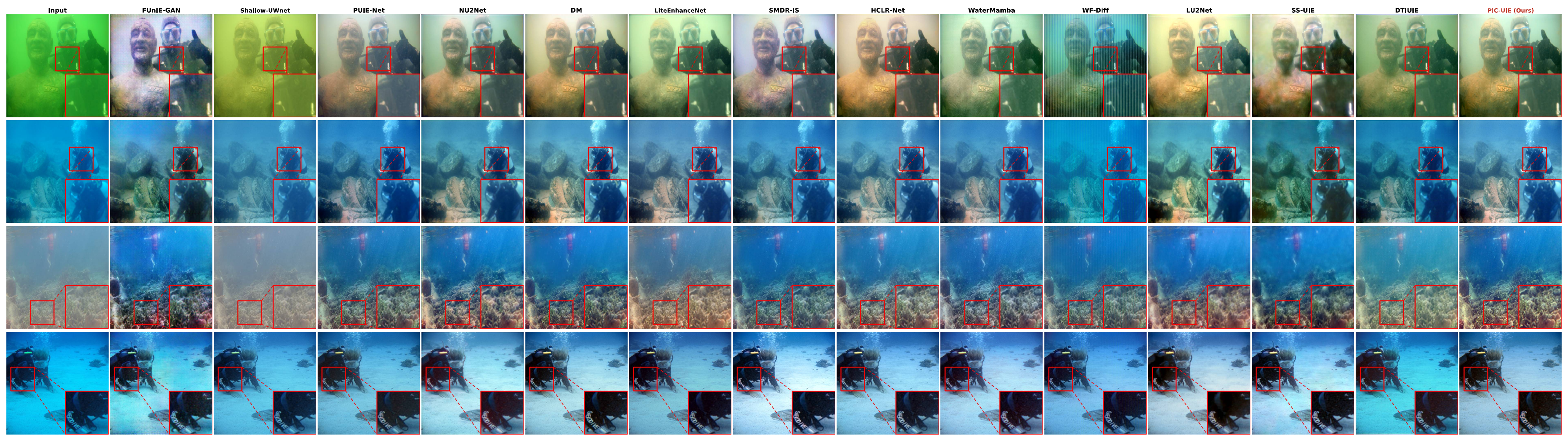}
\caption{Complete zero-shot qualitative comparison on U45.}
\label{fig:supp-u45}
\end{figure*}

\begin{figure*}[!t]
\centering
\includegraphics[width=\textwidth]{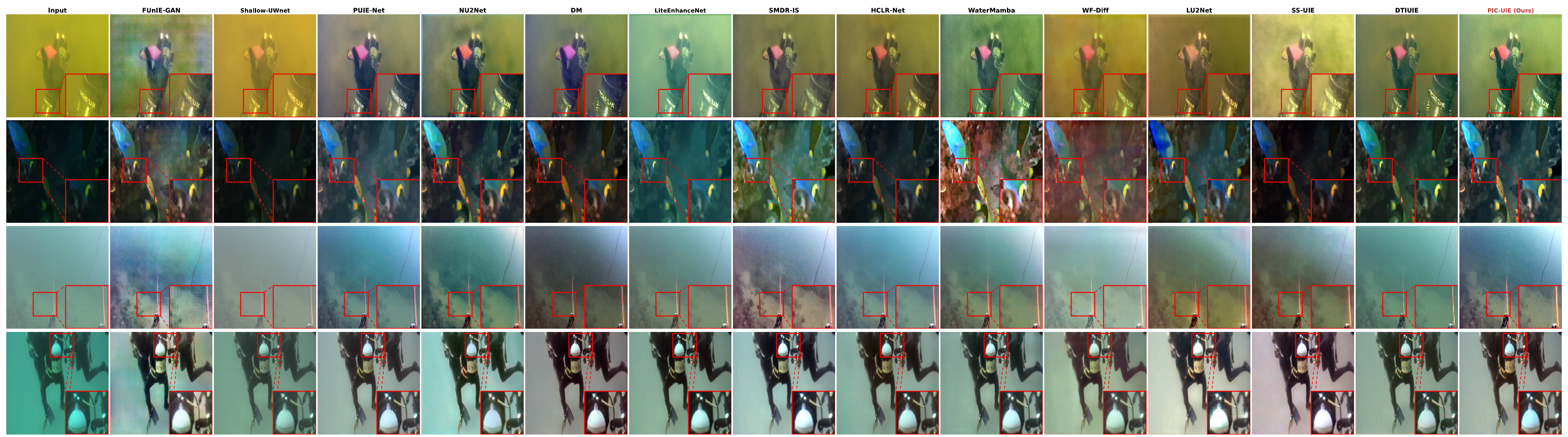}
\caption{Complete zero-shot qualitative comparison on C60.}
\label{fig:supp-c60}
\end{figure*}

\begin{figure*}[!t]
\centering
\includegraphics[width=\textwidth]{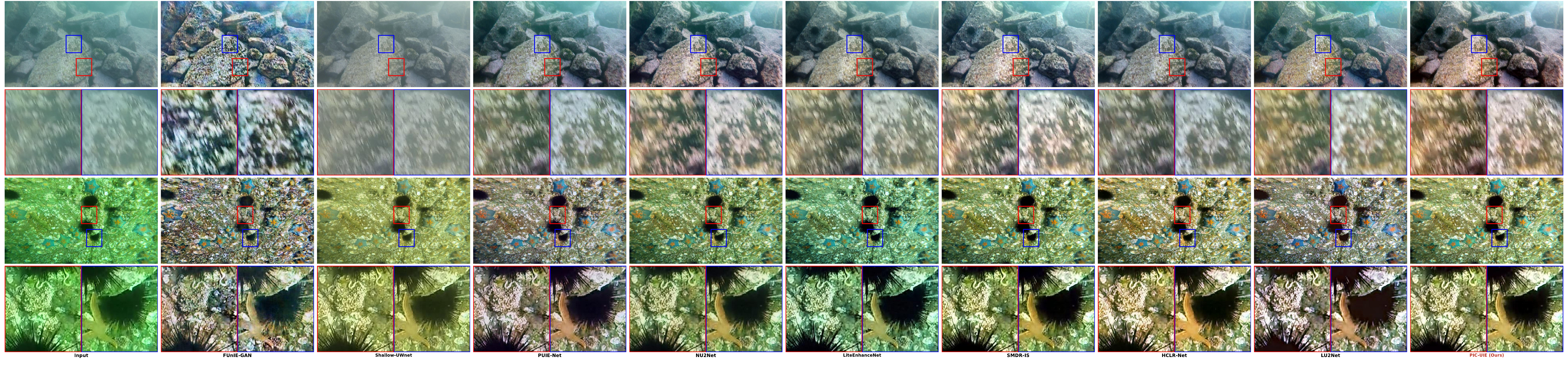}
\caption{Complete native 1080P comparison. Full images and marked crops expose
local color, texture, and artifact differences among methods.}
\label{fig:supp-1080}
\end{figure*}

\begin{figure*}[!t]
\centering
\includegraphics[width=\textwidth]{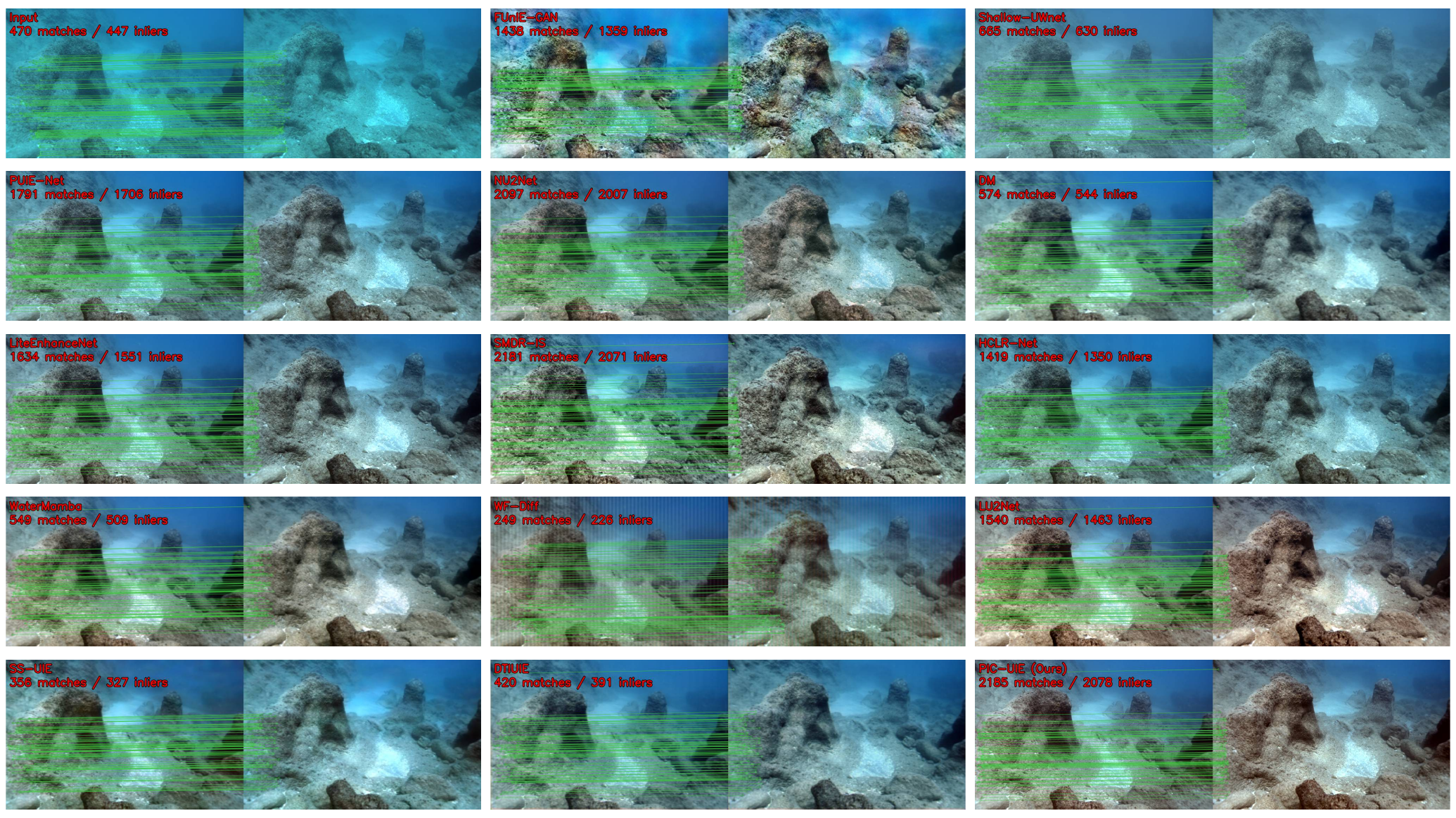}
\caption{FLSea feature-matching examples under the SIFT/RANSAC protocol
described above.}
\label{fig:supp-flsea}
\end{figure*}

\begin{figure*}[!t]
\centering
\includegraphics[width=\textwidth]{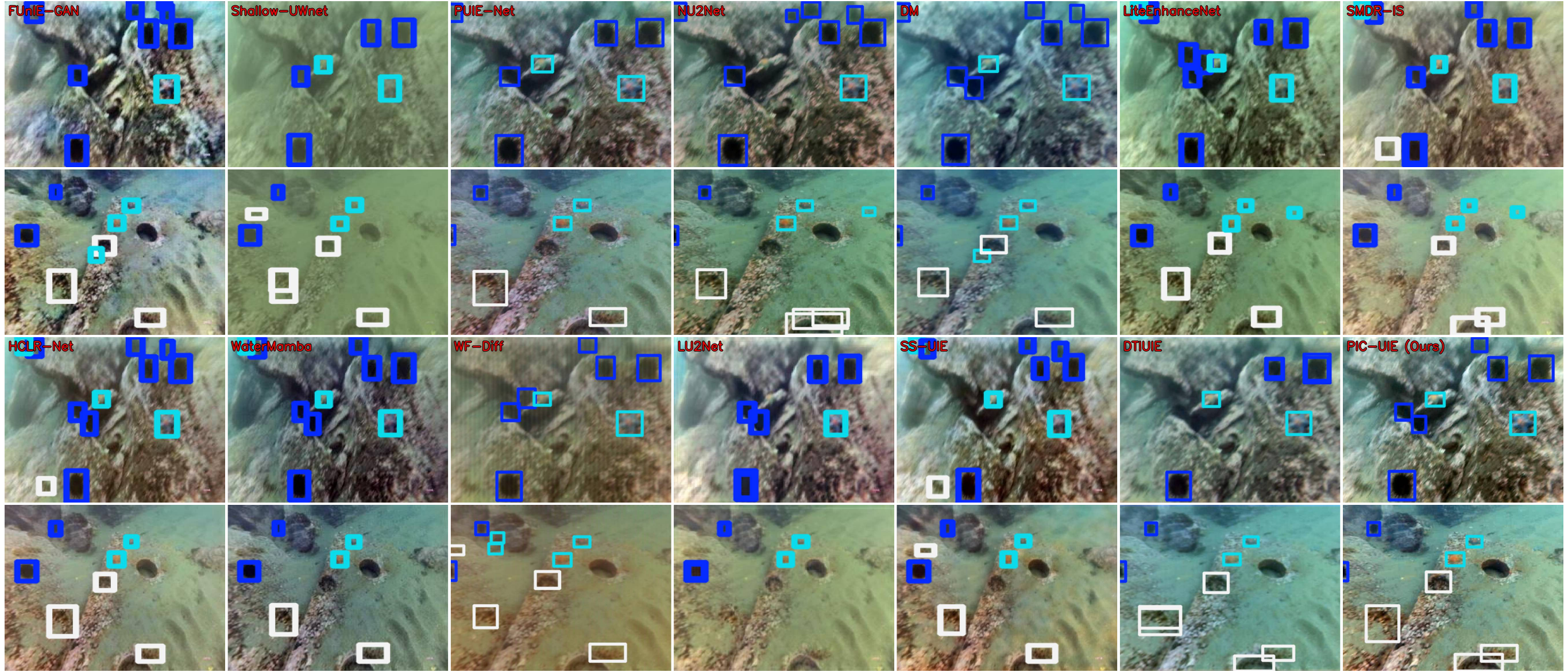}
\caption{Representative URPC2019 detections from the separately trained
YOLO11n models.}
\label{fig:supp-yolo}
\end{figure*}